\pdfoutput=1

\documentclass[11pt]{article}

\usepackage{emnlp2021}

\usepackage{times}
\usepackage{latexsym}

\usepackage[T1]{fontenc}

\usepackage[utf8]{inputenc}

\usepackage{microtype}

\usepackage{soul}
\usepackage{booktabs}
\usepackage{graphicx}
\usepackage{booktabs,multirow}
\usepackage{caption}
\usepackage{subcaption}
\usepackage{lingmacros}

\usepackage{amsmath,amssymb,mathtools,amsthm}
\usepackage{bm}
\usepackage{bbm}
\usepackage{soul}
\usepackage{framed}

\usepackage{textcomp}

\usepackage{url}
\usepackage{xspace}

\DeclareMathAlphabet\mathbfcal{OMS}{cmsy}{b}{n}
\DeclareMathOperator*{\argmax}{arg\,max}

\newcommand{\rrrrc}{$\mathbfcal{R}^4\mathbfcal{C}$\xspace}

\newcommand{\repourl}{\url{https://github.com/StonyBrookNLP/suqa}}
\newcommand{\name}{SuQA}
\newcommand{\baselinename}{Extr}
\newcommand{\xg}{AX}

\newtheorem{definition}{Definition}

\newcommand{\ctext}[3][RGB]{%
  \begingroup
  \definecolor{hlcolor}{#1}{#2}\sethlcolor{hlcolor}%
  \hl{#3}%
  \endgroup
}

\newcommand{\updated}[1]{#1} 

\newcommand{\errorphrase}[1]{\ctext[RGB]{210,210,210}{#1}}
\newcommand{\importantent}[1]{\ctext[RGB]{210,210,210}{#1}}
\newcommand{\goodhi}[1]{\ctext[RGB]{180,220,180}{#1}}
\newcommand{\badhi}[1]{\ctext[RGB]{220,180,180}{#1}}

\title{Summarize-then-Answer: Generating Concise Explanations for Multi-hop Reading Comprehension}

\author{Naoya Inoue$^{\clubsuit,\spadesuit}$, Harsh Trivedi$^{\clubsuit}$, Steven Sinha$^{\clubsuit}$, \\
{\bf Niranjan Balasubramanian}$^{\clubsuit}$, {\bf Kentaro Inui}$^{\diamondsuit,\spadesuit}$ \\
  $\clubsuit$ Stony Brook University, $\spadesuit$ RIKEN \\
  $\diamondsuit$ Tohoku University \\
  \texttt{\{ninoue,hjtrivedi,stsinha,niranjan\}@cs.stonybrook.edu} \\
  \texttt{inui@tohoku.ac.jp} \\
}

\begin{document}
\maketitle
\begin{abstract}

How can we generate concise explanations for multi-hop Reading Comprehension (RC)? The current strategies of identifying supporting sentences can be seen as an extractive question-focused summarization of the input text. However, these extractive explanations are not necessarily concise i.e. not minimally sufficient for answering a question. Instead, we advocate for an abstractive approach, where we propose to generate a question-focused, abstractive summary of input paragraphs and then feed it to an RC system. Given a limited amount of human-annotated abstractive explanations, we train the abstractive explainer in a semi-supervised manner, where we start from the supervised model and then train it further through trial and error maximizing a conciseness-promoted reward function. Our experiments demonstrate that the proposed abstractive explainer can generate more compact explanations than an extractive explainer with limited supervision (only 2k instances) while maintaining sufficiency.

\end{abstract}

\section{Introduction}

\begin{figure}[t!]
\includegraphics[width=\linewidth]{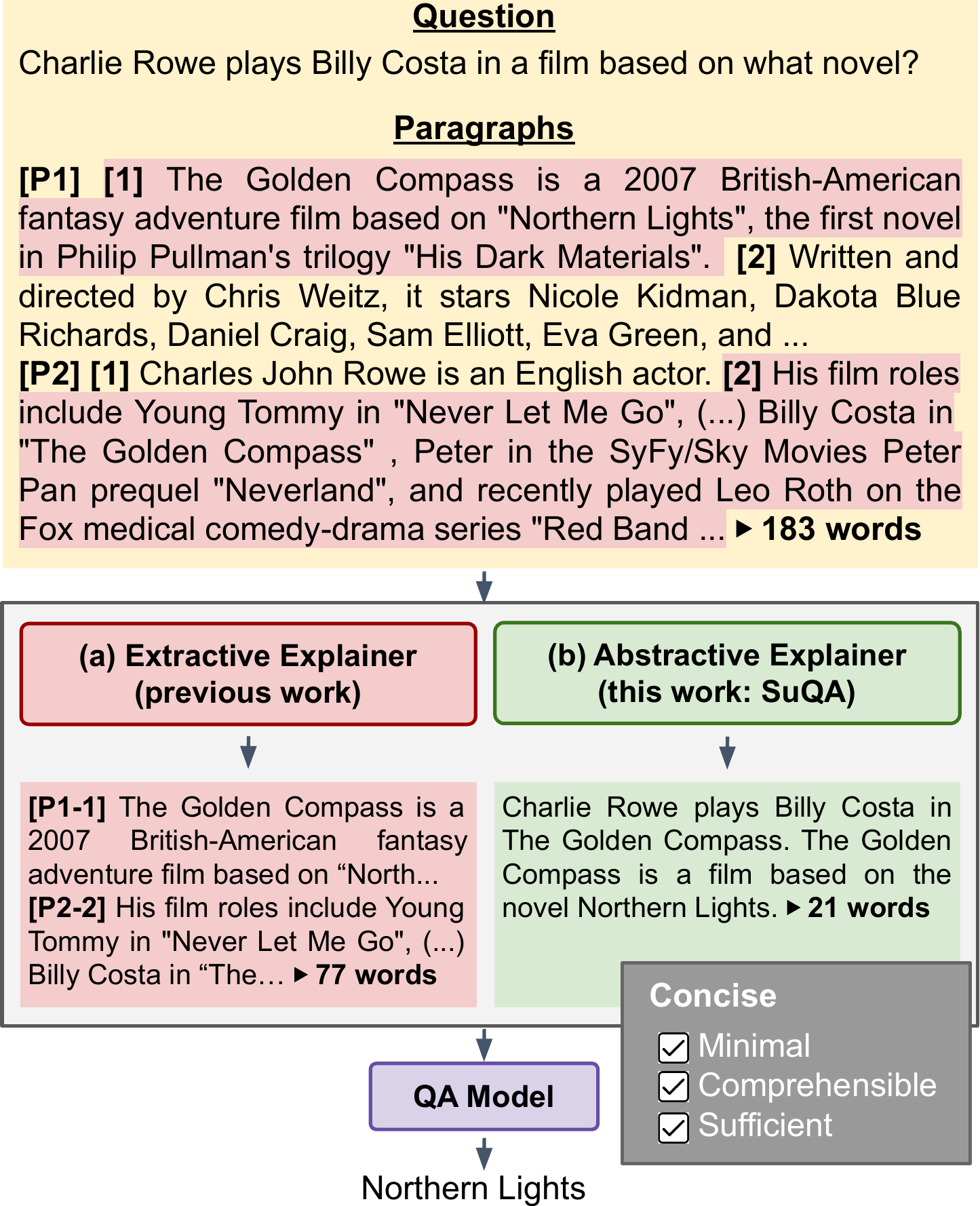}
\caption{
Summary of SUmmarizer-augmented QA (\name{}).
To generate more concise (i.e. minimal, sufficient and comprehensible) explanations, \name{} augments QA module with an abstractive explainer.\footnotemark
}
\label{fig:killer}
\end{figure}

\footnotetext{\updated{
Our implementation is publicly available at \repourl{}.
}}

Recent approaches to multi-hop Reading Comprehension (RC) have greatly improved its \emph{explainability}, models ability to explain their own answers~\cite{thayaparan2020survey}.
Some adopt a pipelined architecture, where they generate an explanation first and then use it to answer the question. This  ``faithful-by-construction'' approach is aimed at ensuring that generated explanations are closer to the systems' internal reasoning (i.e. \emph{faithfulness}).
The explanation generation step is typically formulated as a sentence selection task over the input text --- selecting a set of sentences which provide support for the answer output by the model ~\cite[etc.]{yang-etal-2018-hotpotqa,groeneveld-etal-2020-simple}.

However, the main problem with these approaches is that the explanations obtained from the sentence selection tasks are not always minimal, sufficient, and comprehensible. The extractive explanations can include extraneous or superfluous texts which express information that is not necessary for answering questions. For example, as shown in Fig.~\ref{fig:killer}~(a), the fragments such as \emph{2007 British-American fantasy adventure} and \emph{Young Tommy in ``Never Let Me Go''} are not needed to explain the answer \emph{Northern Lights}. Secondly, the extractive explanations may also not be sufficient: the interpretation of explanations may be dependent on its original paragraphs (e.g. pronouns). In Fig.~\ref{fig:killer}~(a), \emph{His film roles} means \emph{Charles Rowe's film}, but this is not included in the extractive explanation. These types of gaps can also limit comprehensibility of the explanations.

In this work, we target \emph{concise explanations} which provide minimal, sufficient and comprehensible information related to the answer. This can also be seen as targeting an abstractive question-focused summary. To this end, we propose \emph{SUmmarizer-augmented QA (\name{})}, an RC system augmented with an \emph{abstractive explainer} component that generates an abstractive summary of explanations, which is then fed to a a separate QA module to produce an answer. An abstractive explainer can summarize longer sentences into short phrases and replace pronouns with their referent, leading to more compact and sufficient explanations compared to extractive explanations.
For example, as shown in Fig.~\ref{fig:killer}~(b), the abstractive explainer, unlike an extractive one, is allowed to remove unnecessary information such as \emph{2007 British-American fantasy adventure}, and to generate context-independent sentences such as \emph{Charlie Rowe plays Billy Costa in The Golden Compass}, instead of \emph{His film roles includes...}.

%
However, developing such an abstractive explainer imposes a significant challenge because of the limited amount of human-annotated abstractive explanations available and prohibitively high costs in  extending these~\cite{inoue-etal-2020-r4c}.
Given this limited supervision, how can we ensure that generated explanations are sufficient while promoting compression?

Our solution is to teach an abstractive explainer through trial and error maximizing a conciseness-promoting reward function in a reinforcement learning (RL) framework. The reward function assesses generated explanations against various criteria related to conciseness, such as linguistic acceptability, abstractiveness, and the accuracy of RC module's prediction on the generated explanations.
By doing so, the model gradually learns to extract and summarize information from input texts so that they help the RC module arrive at the correct answers. Also, because the explainer aims to produce abstractive summaries, we can initialize the explainer with an abstractive summarizer that is \emph{pretrained} on standard summarization datasets. 

We evaluate the proposed approach on HotpotQA~\cite{yang-etal-2018-hotpotqa}, one of the most popular multi-hop RC datasets.
The findings of this paper can be summarized as follows:
\begin{itemize}
    \item The semi-supervised abstractive explainer can generate more compact and sufficient explanations than extractive explanations while keeping explanations informative for answering questions.
    Compared to extractive ones, the abstractive explanations have a compression rate that is $\times 2.9$ higher, and improve human-judged sufficiency by 2.5 points, without incurring any significant drop in the QA accuracy.
    
    \item Even small amounts of human-annotated explanation supervision significantly improve the conciseness of generated explanations. 
    For example, incorporating even 298 instances of annotated explanations makes the compression rate  $\times 1.3$ higher and improves human-judged sufficiency by $+11.0$ points compared to the setting with no supervision for explanations.
\end{itemize}

\section{Related work}

\paragraph{Explainable NLP}
Three aspects of explainability have been explored~\cite{jacovi-goldberg-2020-towards}: (i) comprehensibility to humans~\cite{NEURIPS2018_4c7a167b,rajani-etal-2019-explain}, (ii) faithfulness, correlation with systems' internal decision~\cite{kumar-talukdar-2020-nile,glockner-etal-2020-think}, (iii) conciseness, namely minimality, comprehensibility and sufficiency for solving an end task~\cite{paranjape-etal-2020-information}.

Earlier approaches to explainable NLP focus on comprehensibility~\cite{NEURIPS2018_4c7a167b,rajani-etal-2019-explain}, and then the community moves towards ensuring faithfulness by a system's architecture (\emph{faithful by construction}), ranging from Natural Language Inference~\cite{,kumar-talukdar-2020-nile}, Fact Verification~\cite{glockner-etal-2020-think} to Question Answering~\cite{latcinnik2020explaining,groeneveld-etal-2020-simple,yadav-etal-2020-unsupervised}.

Conciseness, in contrast, has been relatively unexplored.
One exception is \newcite{paranjape-etal-2020-information}, who propose to learn to extract a minimal set of input sentences that are useful for solving downstream tasks by imposing information bottleneck on the NLP framework.
Although our work shares the similar spirit with their work, unlike our work, their explainer is extractive.
Our work is the first to incorporate abstractive explainers into RC systems.

To date, more NLP datasets are being annotated with explanations~\cite{wiegreffe2021teach}, but most of them are based on extractive explanations~\cite[etc.]{yang-etal-2018-hotpotqa,deyoung-etal-2020-eraser}.
For abstractive explanations, there are a few resources: textual entailment dataset~\cite{NEURIPS2018_4c7a167b}, and question answering dataset in non-RC settings (i.e. input paragraphs are not given)~\cite{Jansen2018,rajani-etal-2019-explain}.
As for RC, \newcite{inoue-etal-2020-r4c} annotate HotpotQA~\cite{yang-etal-2018-hotpotqa} with abstractive explanations, but only 2k of them (i.e. 3\% of the whole dataset) are annotated.

\paragraph{Abstractive explainer}
\updated{
A similar pipeline model has been proposed for textual entailment~\cite{NEURIPS2018_4c7a167b} and commonsense QA~\cite{rajani-etal-2019-explain}, where the model first generates an explanation, and then the downstream classifier consumes it to predict a task label.
Although the architecture is the same as ours, the training process is different: they train the explainer in a fully supervised manner using input-explanation pairs, while our work additionally leverages a signal from the downstream QA model in RL.
As demonstrated in \S\ref{sec:results}, we show that this additional training is crucial when few annotated explanations are available.
}

\updated{
Generating abstractive explanations is closely related to query-focused summarization (QFS), where a few datasets are publicly available~\cite{dang-2006-duc,baumel2016tcduc,nema-etal-2017-diversity,pasunuru2021data}.
However, the task setting of QFS is radically different from our problem setting, which makes it difficult to leverage the datasets and models in a straightforward manner.
The QFS task typically consists of non-question queries (e.g. keywords or complex sentences) or opinion-oriented questions (e.g. \emph{Is X a good idea?}), and gold summaries are not guaranteed to contain all information required for answering questions.
We leave it the future work to explore how to effectively use their datasets and models in our task.
}

\section{\name{}: SUmmarizer-augmented QA}
\label{sec:proposed}

\begin{figure*}[t]
\centering
\includegraphics[width=0.9\linewidth]{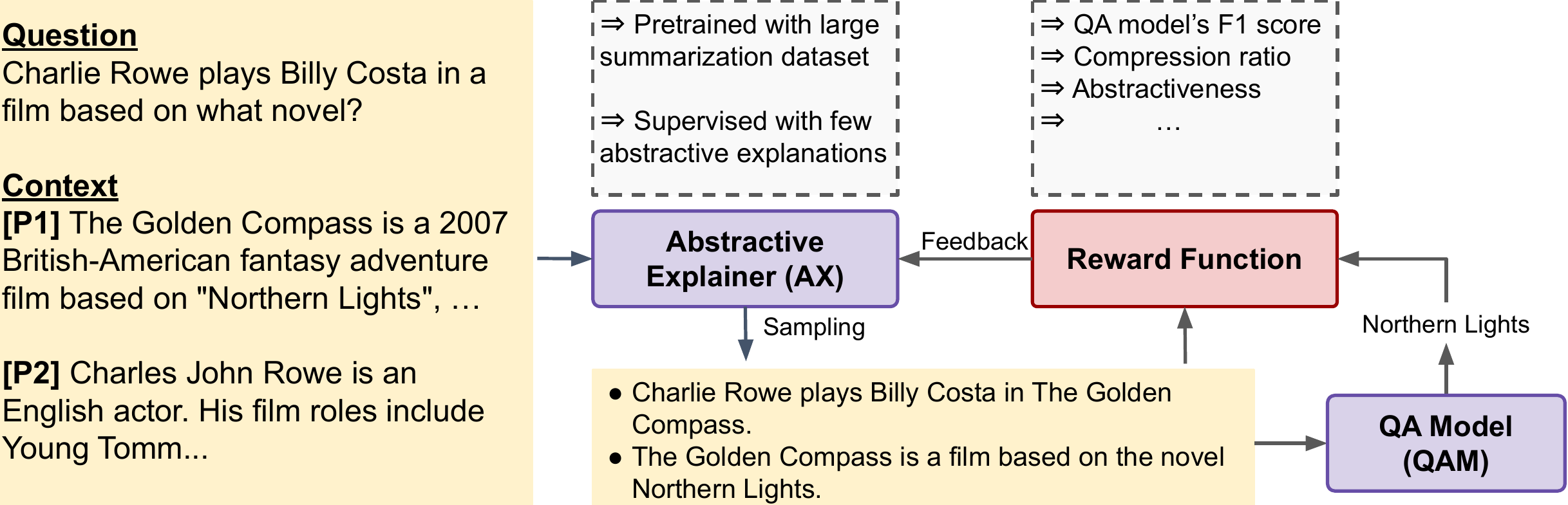}
\caption{
Training regime of the proposed method.
We pretrain the \xg{} with a large summarization dataset and finetune it on a limited amount of human-annotated explanations (\S\ref{sec:pretrain}).
We then train it further through indirect supervision from the QAM using Reinforcement Learning (\S\ref{sec:rl}).
}
\label{fig:architecture}
\end{figure*}

\updated{
Extractive explanations may contain superfluous information that is not necessary for answering questions or may not be sufficient for answering questions.
We address this issue by generating concise explanations defined as follows.

\begin{definition}
An explanation is \emph{concise} if it is (i) minimal, (ii) comprehensible, and (iii) sufficient for answering the question.
\end{definition}
}

Fig.~\ref{fig:killer} summarizes the overall architecture.
To ensure the faithfulness of explanations, we use a pipeline architecture consisting of two main components: (i) an \emph{abstractive explainer} (\xg{}) and (ii) \emph{QA module} (QAM) (\S\ref{sec:components}).
The \xg{} takes a question and paragraph as inputs and is responsible for generating a question-focused, abstractive summary of input paragraphs.
The QAM then answers the question solely based on the generated summary.
This summary is supposed to contain information necessary for answering questions and is the only factor that the QAM relies on.
Thus, the generated summary can be interpreted as a faithful explanation of the model.

\subsection{Architecture}
\label{sec:components}

First, we formalize the overall pipeline.
Given a question $q$ and paragraphs $p$, we first generate the most-likely explanation $e$ as follows:
\begin{equation}
    e = \argmax_{e'} p_\pi(e'|q, p),
\end{equation}
where $p_\pi$ is the \xg{}.
We then answer the question $q$ \emph{solely} based on the generated explanation $e$:
\begin{equation}
    a = \argmax_{a'} p_\phi(a'|q, e), \label{eq:answerprob}
\end{equation}
where $p_\phi$ is the QAM.
Our architecture is agnostic to the implementation of \xg{} and QAM as long as they are differentiable.

From the viewpoint of probabilistic models, this formulation is a special case of a probabilistic latent variable model of $p(a|q, p)$ where explanations are treated as latent variables, similar to retrieval-augmented language models~\cite{pmlr-v119-guu20a,NEURIPS2020_6b493230}.
Specifically, we have $p(a|q, p) = \sum_{e} p_\phi(a|q, e) p_\pi(e|q, p)$, assuming $p_\phi(a|q, e, p) = p_\phi(a|q, e)$.
Replacing the sum with $\argmax$ yields Equation~\ref{eq:answerprob}.
The main challenge is that $p_\pi(e|q, p)$ is not a retriever but a text generator.

\paragraph{Abstractive explainer (\xg{})}
It takes a paragraph $p$ and a question $q$ as an input, and outputs an explanation $e$.
We implement the \xg{} using a sequence-to-sequence generation model as follows:
\begin{equation}
    p_\pi(e|q, p) = \prod_t^n p_\pi(e_t|e_{<t}, q, p)
\end{equation}
In our experiments, we use BART~\cite{lewis-etal-2020-bart}.
We simply concatenate $q$ and $p$ into one text with a separator token to generate a question-focused summary of the paragraph.

\paragraph{QA module (QAM)}
It takes a question $q$ an explanation $e$ generated by the \xg{} as an input, and outputs an answer $a$.
We implement the QAM as a generation-based question answering module.
\begin{equation}
    p_\phi(a|q, e) = \prod_t^n p_\phi(a_t|a_{<t}, q, e)
\end{equation}

\section{Training}
\label{sec:training}

Fig.~\ref{fig:architecture} shows an overview of our training regime.
The main challenge of training the \xg{} is that human-annotated explanations are rarely available for question-answer pairs, though the conciseness of explanations heavily relies on human judgement.
To address this issue, we train the \xg{} in a semi-supervised manner.

\subsection{Supervised training with summarization and explanation generation}
\label{sec:pretrain}

\updated{
Because the \xg{} aims to produce abstractive summaries, we initialize the \xg{} with an abstractive summarizer that is pretrained on standard summarization datasets.
As we will see later (\S\ref{sec:training_str}), this initialization is one of the key ingredients for the \xg{}.
}

Given a training dataset consisting of QA pairs annotated with its gold explanations, we train the \xg{} with a standard teacher forcing approach.
Specifically, we minimize the following loss:
\begin{equation}
    L_{\mathrm{ML}} = \sum_{t=1}^n \log p_\pi(y^*_t|y^*_{<t}, q), \label{eq:lml}
\end{equation}
where $q$ is a question, and $(y^*_1, y^*_2, ..., y^*_n)$ is a human-annotated explanation for the QA pair.

\subsection{Semi-supervised training}
\label{sec:rl}

Although the fully supervised training provides the \xg{} with direct signals, large-scale annotation of such abstractive explanation is prohibitively costly~\cite{inoue-etal-2020-r4c}.
Thus, after training the \xg{} in a supervised fashion, we further train the \xg{} through indirect supervision from answers, which are much cheaper to annotate.

We use the RL framework and design a reward function that assesses the goodness of generated explanations based on answers and sentence-level supporting facts. A state here is a sequence of explanation tokens generated so far $y_{<t}$, an action is to generate a token, and the policy function is a probability distribution $p_\pi(y_t|y_{<t}, q)$ of tokens given by the \xg{}, as with previous work on RL-based language generation~\cite[etc.]{Rennie_2017_CVPR}.
Given a reward function $r(\cdot)$ which we describe later, we optimize the policy function $p_\pi(y_t|y_{<t}, q)$ via self-critical training~\cite{Rennie_2017_CVPR} as follows:
\begin{equation}
    L_{\mathrm{RL}} = -\frac{1}{n} \sum_{t=1}^n (r(y') - r(\hat{y})) \log p_\pi(y'_t|y'_{<t}, q),
\end{equation}
where $y'$ is a sampled explanation according to the current policy, and $\hat{y}$ is an explanation generated by a greedy decoding.
\updated{
$r(\hat{y})$ is called a baseline reward that stabilizes the training process by reducing the variance in the gradient.
}
To prevent generated explanations from deviating too much from gold explanations, we jointly optimize the RL loss with the supervised loss: our final loss is $L_{\mathrm{RL}} + \lambda L_{\mathrm{ML}}$, where $\lambda$ is a weight of the ML loss.
In our experiments, we used $\lambda = 0.1$.

\subsection{Reward function}
\label{sec:reward}

Given question $q$, input paragraphs $c$, and explanation $e$, we define the reward function as a geometric mean of $N$ elemental reward functions:
\begin{equation}
    r(e) = \mathrm{gmean}(\{ r_i(q, c, e) \}_{i=1}^N)
\end{equation}
\updated{
The intuition here is that we combine elemental reward functions with ``AND'' operator: if one of elemental reward functions gives zero, the explanation must not be rewarded.
}
We introduce three types of elemental reward functions as follows.

\paragraph{Summarization rewards} promote the \xg{} to generate more compact summaries.
To keep the summary relevant to the question, we also incorporate the relevance of generated explanations to input paragraphs and questions.
Let $P,Q$ be a set of tokens, and the $P$'s coverage of $Q$ be $\mathrm{cov}(P, Q) = |P \cap Q|/|Q|$.
Let $\mathrm{ng}(X, i)$ be a set of $i$-grams in $X$, and $w(X)=\mathrm{ng}(X, 1)$.
\begin{itemize}
 	\item Compression ratio of $e$ w.r.t. input paragraphs: $1 - (\text{\# tokens in } e / \text{\# tokens in } c)$
 	\item Abstractiveness of $e$ w.r.t. input paragraphs: $1/4 \sum_i^4 (1 - \mathrm{cov}(\mathrm{ng}(c, i), \mathrm{ng}(e, i)))$.
    \item Relevance of $e$ to input paragraphs based on unigrams: $\mathrm{cov}(w(c), w(e))$
 	\item $e$'s coverage of question: $\mathrm{cov}(w(e), w(q))$
\end{itemize}

\paragraph{Sufficiency rewards} ensure that generated explanations are sufficient, i.e. useful for answering questions.
\begin{itemize}
	\item F1 score of the QAM's predicted answer: we feed $e$ into the QAM and calculate the answer F1 score of the predicted answer.
    \item Existence of gold answer span: $1$ if $e$ contains the gold answer span; $0$ otherwise.
\end{itemize}

\paragraph{Comprehensibility rewards} ensure the comprehensibility of generated explanations to humans.
\begin{itemize}
    \item Linguistic acceptability: we feed $e$ into a pretrained CoLA~\cite{warstadt2018neural} scorer.
    In our experiments, we use RoBERTa-base finetuned on the CoLA dataset.\footnote{\url{https://huggingface.co/textattack/roberta-base-CoLA}}

    \item Sampling noisiness: $1$ if $\log p_\pi(e|q,p) > T$; $0$ otherwise. This is to prevent noisy explanations from being rewarded. We use $T = -50$.
    \item Well-formedness: $1$ if $e$ has repetition or too long words, starts from pronouns, or ends without period; $0$ otherwise.
\end{itemize}

\section{Evaluation}

\subsection{Dataset}

We use HotpotQA~\cite{yang-etal-2018-hotpotqa}, which consists of 90,564 training and 7,405 development instances.\footnote{\url{https://hotpotqa.github.io/}}
All instances are annotated with extractive explanations called \emph{supporting facts, or SFs}, sentences that are required to answer questions from input documents.
We use the distractor setting in our experiments.

For human-annotated explanations, we use \rrrrc~\cite{inoue-etal-2020-r4c},\footnote{\url{http://naoya-i.github.io/r4c}}
which annotates 2,379 training instances (3\% of the training instances) and 2,541 development instances from HotpotQA with reasoning steps.
The reasoning steps are abstractive explanations that describe information necessary for deriving answers, consisting of entity relation triplets in natural language texts (e.g. \emph{(Biden, is a president of, US)}).
We concatenate entities and its relation into one sentence for training the \xg{}.

\subsection{Relevant paragraph prediction}
\label{sec:parapred}

To select relevant paragraphs for the \xg{}, we trained a ranker that ranks paragraphs according to its relevance to questions.
The ranker takes a question and one paragraph as an input and outputs a relevance score.
To train the ranker, we used a binary cross entropy loss, where paragraphs containing gold SFs (henceforth, \emph{supporting paragraphs}) are used as positive instances and the other distractor paragraphs are negative instances.
Following~\newcite{kim-etal-2020-dense}, we also randomly sample one supporting paragraph from other questions for each question and used them as negative instances.

At test time, we retain top-$k$ paragraphs and give them to the \xg{}.
We use $k=3$ because HotpotQA has two supporting paragraphs always.
Our evaluation shows that all supporting paragraphs are included at top-$k$ ranked paragraphs in 97.4\% of dev instances on HotpotQA.
When training the \xg{}, we gave gold supporting paragraphs and randomly selected distractor paragraphs to the \xg{}.
To implement the ranker, we use a standard sequence classifier on top of RoBERTa-large~\cite{roberta2019}.

\subsection{Setup}
\label{sec:setup}

\paragraph{Models}
We create \emph{\baselinename{}}, a simple baseline model that resembles a typical extraction-based explainable NLP architecture~\cite{glockner-etal-2020-think,paranjape-etal-2020-information}.
Here, we train the \xg{} using Eq.~\ref{eq:lml} only, where we use SFs as supervision.

We denote our proposed model as \emph{\name{}}.
To see the effectiveness of RL, we have \emph{\name{}-NoRL}, a model trained with annotated explanations using Eq.~(\ref{eq:lml}) \emph{without additional RL training}.
\name{}-NoRL resembles fully-supervised, generation-based explain-then-predict models by \newcite{NEURIPS2018_4c7a167b,rajani-etal-2019-explain}.

\paragraph{\xg{}}
We initialize the \xg{} with DistilBART finetuned on CNN/Daily Mail, one of large, standard datasets of summarization~\cite{shleifer2020pretrained}.
During training, we feed supporting paragraphs as an input to the model.
At test time, we use predicted relevant paragraphs from \S\ref{sec:parapred} as an input.
For hyperparameter tuning, we reserve 500 training instances as a validation dataset.
See \S\ref{sec:training detail} in Appendix for further details.

\paragraph{QAM}
\updated{
We use UnifiedQA-base~\cite{khashabi-etal-2020-unifiedqa} as the QAM and freezed it during training.
Ideally, the \xg{} should learn from a ``perfect'' QA model that does not perform disconnected reasoning~\cite{trivedi-etal-2020-multihop}.
However, such a QA model is not available at the moment.
We thus simulate it by using UnifiedQA~\cite{khashabi-etal-2020-unifiedqa}, a T5~\cite{2020t5}-based QA model finetuned on a diverse set of QA datasets (e.g. SQuAD, NarrativeQA, RACE) \emph{excluding} HotpotQA.
We expect this to discourage the QAM from giving correct answers for insufficient explanations by disconnected reasoning, which improves the quality of reward function of RL.
At test time, we use UnifiedQA finetuned on HotpotQA, whose performance is shown in Table~\ref{tab:main} (see QAM w/o \xg{}).
}

\subsection{Evaluation measures}
\label{sec:evalm}

\paragraph{Conciseness} To assess the \emph{compactness} of generated explanations, we calculate (i) a compression ratio (\emph{Cm}), \# tokens in an input paragraph divided by \# tokens in a generated explanation, and (ii) abstractiveness (\emph{Abs}) with respect to a given paragraphs selected by the paragraph ranker, calculated by the equation from \S\ref{sec:reward}.

To assess the \emph{sufficiency} of generated explanations, we use crowdsourcing.
Given a generated explanation and its original question, five crowdworkers are asked to judge if generated explanations alone provide sufficient information for answering the question in a 3-point Likert scale (yes, likely, no) plus ``unsure''.
To reliably estimate the quality of explanations, we additionally ask them answers that they inferred from the given explanations.

To aggregate each annotator's judgement, we first replace crowdworker's submission with `no' when (i) the answer is different from the gold standard answer, or (ii) the judgement is unsure, and replace `likely' with `yes'.
We then used MACE~\cite{hovy-etal-2013-learning} to aggregate all the judgements (\emph{Suf}).
Due to the cost,\footnote{We paid the workers \$9/hr.} we evaluate 100 gold explanations and 200 generated explanations for each configuration.
We obtained Krippendorff's $\alpha$ of 0.298 on average, indicating a fair agreement.
See \S\ref{sec:humaneval} in Appendix for further details of crowdsourced judgement.

In some experiments, we report the similarity between generated explanations and human-annotated explanations as a proxy for sufficiency, due to the cost of human evaluation.
We employ ROUGE-2~\cite{Lin2004} (\emph{RG2}), which is proven a high correlation between human ratings on several summarization datasets~\cite{bhandari-etal-2020-evaluating}.

\paragraph{QA performance} We report \emph{F1}, one of the official evaluation measures of HotpotQA.

Given that our ultimate goal is to create an explainable RC system, we also introduce \emph{XF1}, new evaluation measure:
\begin{eqnarray}
    \mathrm{XF1} = \frac{1}{N} \sum_i^N \mathrm{suf}(i) \cdot \mathrm{F1}(i),
\end{eqnarray}
where $N$ is the number of instances in the dataset, $\mathrm{suf}(i)$ is a crowdsourced sufficiency label (yes=1, no=0), and $\mathrm{F1}(i)$ is a F1 score of $i$-th instance.
This captures how well the system generates sufficient explanations \emph{and} predicts the correct answer.

\subsection{Results and discussion}
\label{sec:results}

\paragraph{Abstractive explanations are more concise (i.e. compact and sufficient) than extractive ones.}
To understand the advantage of abstractive explanations, we compare gold extractive explanations (Gold SF) with gold abstractive explanations (Gold XP) in Table~\ref{tab:upper_bound}.
It clearly indicates that abstractive explanations are more abstract and compact than extractive ones.
Surprisingly, it also shows that extractive explanations are much less sufficient than abstractive ones.
Our manual inspection of insufficient explanations reveals that 100\% of the explanations do contain gold answer spans, but the interpretation of them depends on the context of input paragraphs that is not included in the explanations (e.g. pronoun referents).
On the one hand, pronouns in abstractive explanations can be replaced with the actual referent, which allows explanations to be more self-contained and compressed.
F1 also improved given more sufficient explanations.

\begin{table}[t]
\centering
\small
\begin{tabular}{lrrrrrr}\toprule
\textbf{Input} &\textbf{Abs} &\textbf{Cm} &\textbf{Suf}$^{\ddagger}$ &\textbf{F1} \\\midrule
Gold SF$^\dagger$ &1.1 &4.4 &72.0 &79.7 \\
Gold SF &1.2 &4.3 &68.0 &74.9 \\
Gold XP$^\dagger$ &\textbf{51.0} &\textbf{11.1} &\textbf{90.0} & \textbf{85.2} \\
\bottomrule
\end{tabular}
\caption{Upper bound study on HotpotQA (HQ) dev set. $\dagger$: evaluated only on 2,541 dev instances annotated with explanations. $\ddagger$: manually evaluated on 100 instances.}
\label{tab:upper_bound}
\end{table}

\begin{table}[t]
\centering
\small
\begin{tabular}{lrrrrrr}\toprule
\textbf{Model} &\textbf{Abs} &\textbf{Cm} &\textbf{Suf}$^\dagger$ &\textbf{F1} & \textbf{XF1}$^\dagger$ \\
\midrule
QAM w/o \xg{} & 0.0 & 1.0 & - & 64.2 & - \\
\baselinename{} (baseline) &0.3 &4.2 &70.0 &\textbf{69.4} &60.5 \\
\name{}-NoRL &40.1 &11.2 &71.5 &65.6 & 62.6 \\
\name{} &\textbf{42.6} &\textbf{12.2} &\textbf{72.5} &67.6 &\textbf{63.7} \\
\bottomrule
\end{tabular}
\caption{Main results on HotpotQA dev set. $\dagger$: evaluated on 200 instances with human-judged sufficiency.}
\label{tab:main}
\end{table}

\begin{table}[t]
\centering
\small

\begin{tabular}{c}
\begin{minipage}{0.5\linewidth}
\begin{tabular}{lrrr}\toprule
&Correct &Wrong \\\midrule
Suf. &\goodhi{124} &16 \\
Insuf. & \badhi{27} &33 \\
\midrule
Total &151 &49 \\
\bottomrule
\end{tabular}
\subcaption{\baselinename{} (baseline)}
\end{minipage}

\begin{minipage}{0.5\linewidth}
\begin{tabular}{lrrr}\toprule
&Correct &Wrong \\\midrule
Suf. &\goodhi{128} &17 \\
Insuf. &\badhi{17} &38 \\
\midrule
Total &145 &55 \\
\bottomrule
\end{tabular}
\subcaption{\name{}}
\end{minipage}
\end{tabular}

\caption{
Sufficiency-Answer correctness matrix.
\name{} gets more correct answers with sufficient explanations (128/145=88\%) than \baselinename{} (124/151=82\%).
}
\label{tab:conf}
\end{table}

\begin{table*}[t]
\centering
\small
\begin{tabular}{p{60mm}p{70mm}p{20mm}}
\toprule
Question & Generated explanation &Gold answer \\\midrule
(a) Who was born first \importantent{Burton Cummings} or \importantent{Sharleen Spiteri}? & \importantent{Burton Cummings} is born on December 31, 1947. \importantent{Sharleen Spiteri} is born on 7 November 1967. &Burton Lorne Cummings \\
(b) The Livesey Hal War Memorial commemorates the fallen of \importantent{which war}, that had over 60 million casualties? &Livesey Hall War Memorial commemorates the fallen of \importantent{World War II}.
\importantent{World War II} had over 60 million casualties. &World War II \\
(c) Charles Barton "Chuck" Kendall, Jr. was reportedly interested in purchasing the Los Angeles Clippers from \importantent{which Jewish-American businessman}? &Charles Kendall, Jr. was reportedly interested in purchasing the Los Angeles Clippers from owner \importantent{Donald Sterling}.
\importantent{Donald Sterling} is a Jewish-American businessman. &Donald Sterling \\
\bottomrule
\end{tabular}
\caption{
Sufficient explanations from \name{}. 
Important entities are gray-highlighted by the author.
}
\label{tab:nice}
\end{table*}

\paragraph{The abstractive explainer generates more concise explanations.}
Now we turn to the proposed models.
The results are shown in Table~\ref{tab:main}.
As consistent with Table~\ref{tab:upper_bound}, it shows that \name{} generates more abstractive, compact and sufficient explanations than the extractive baseline model.
Examples of sufficient explanations generated by \name{} are shown in Table~\ref{tab:nice} (see \S\ref{sec:fulloutputs} in Appendix for more outputs with full input paragraphs).
It shows that the abstractive explainer successfully captures information about important entities in question (e.g. bridging entity \emph{World War II} in (b)).

\updated{
One may think why F1 of \name{} is lower than that of the extractive baseline (-1.8 point) given more sufficient and compressed explanations, which is inconsistent with Table~\ref{tab:upper_bound}.
To obtain further insights, we investigated the relation between the sufficiency of explanations and the correctness of answers in Table~\ref{tab:conf}, where ``Correct'' here means the number of instances with $>0.5$ Answer F1.

Table~\ref{tab:conf} shows that the extractive baseline got 27 correct answers \emph{even when explanations are insufficient} (27/151=17.9\%), while \name{} got 17 correct answers for insufficient explanations (17/145=11.7\%).
This suggests that that the QA module relies on task-unrelated lexical cues --- so-called disconnected reasoning~\cite{trivedi-etal-2020-multihop}, and such task-unrelated cues become unavailable in \name{}'s more compressed explanations, which undesirably degrades the QA performance.
We also experimented with SAE-large~\cite{tu2020sae}, one of the strong QA models in HotpotQA, but got a similar trend.
See \S\ref{sec:largeqa} in Appendix for further details.
We believe that QA performance will improve if one can successfully develop a QA model that performs less shortcut reasoning, which is an emerging research topic in the QA community.
}

\paragraph{The proposed model generates more correct answers with sufficient explanations.}

Our ultimate goal is to predict correct answers \emph{and} to genereate sufficient explanations.
Here we investigate how many instances we generate sufficient explanations \emph{and} predict the correct answer for.
Table~\ref{tab:conf} show that \name{} gets more correct answers with sufficient explanations (128/145=88\%) than the extractive baseline (124/151=82\%).
XF1 in Table~\ref{tab:main} reflects this tendency and now tells a different story from conventional F1: the extractive baseline is now behind the proposed model.

\paragraph{RL helps generate concise explanations.}
As described in \S\ref{sec:setup}, we pretrain the \xg{} with explanations before applying RL.
How much does the additional RL help the \xg{} generate more concise explanations?
The results are shown in Table~\ref{tab:main} (\name{}-NoRL v.s. \name{}).
It indicates that RL is important to obtain more concice explanations in all the aspects of conciseness.

\begin{figure}[t]
\includegraphics[width=\linewidth]{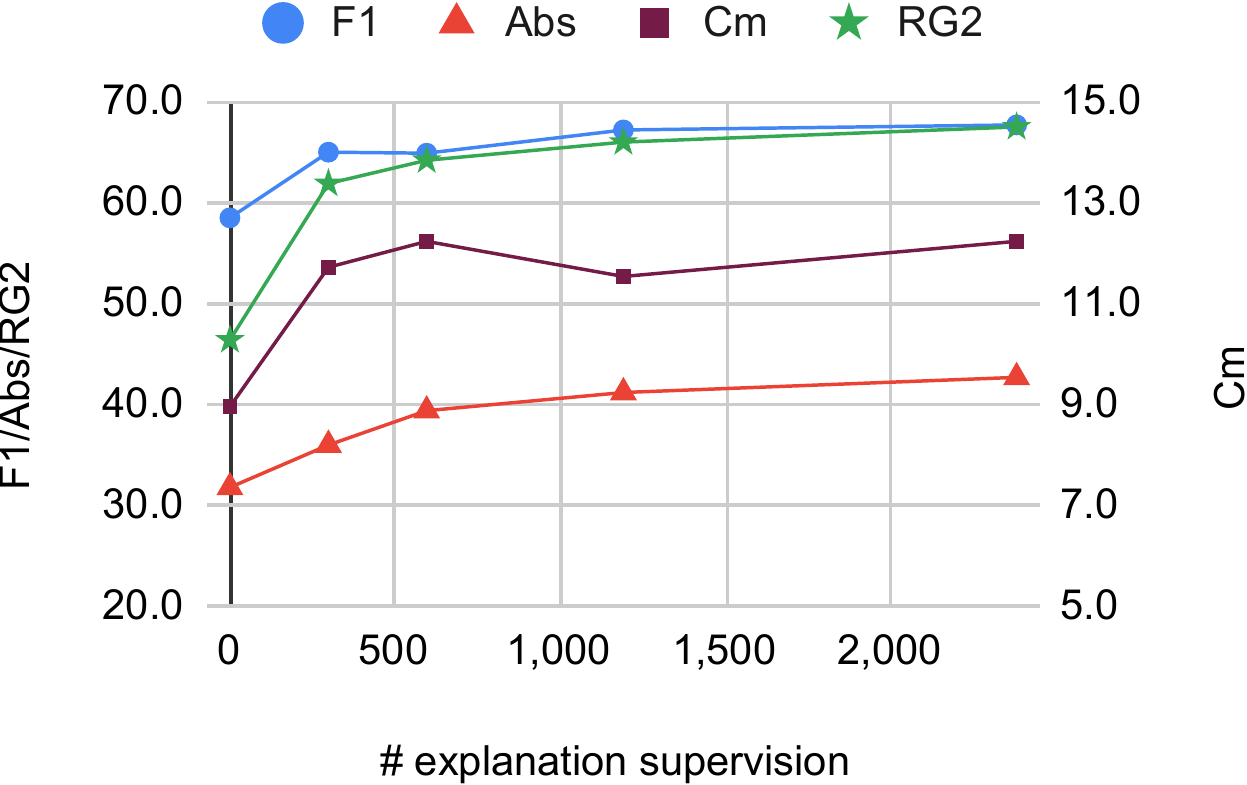}

\caption{
Effect of size of explanation supervision.
Our human-judged sufficiency shows 55.0 at size 0 and 66.0 at size 298, indicating the importance of explanation supervision.
}
\label{fig:size_expl}
\end{figure}

\begin{figure}[t]
\includegraphics[width=\linewidth]{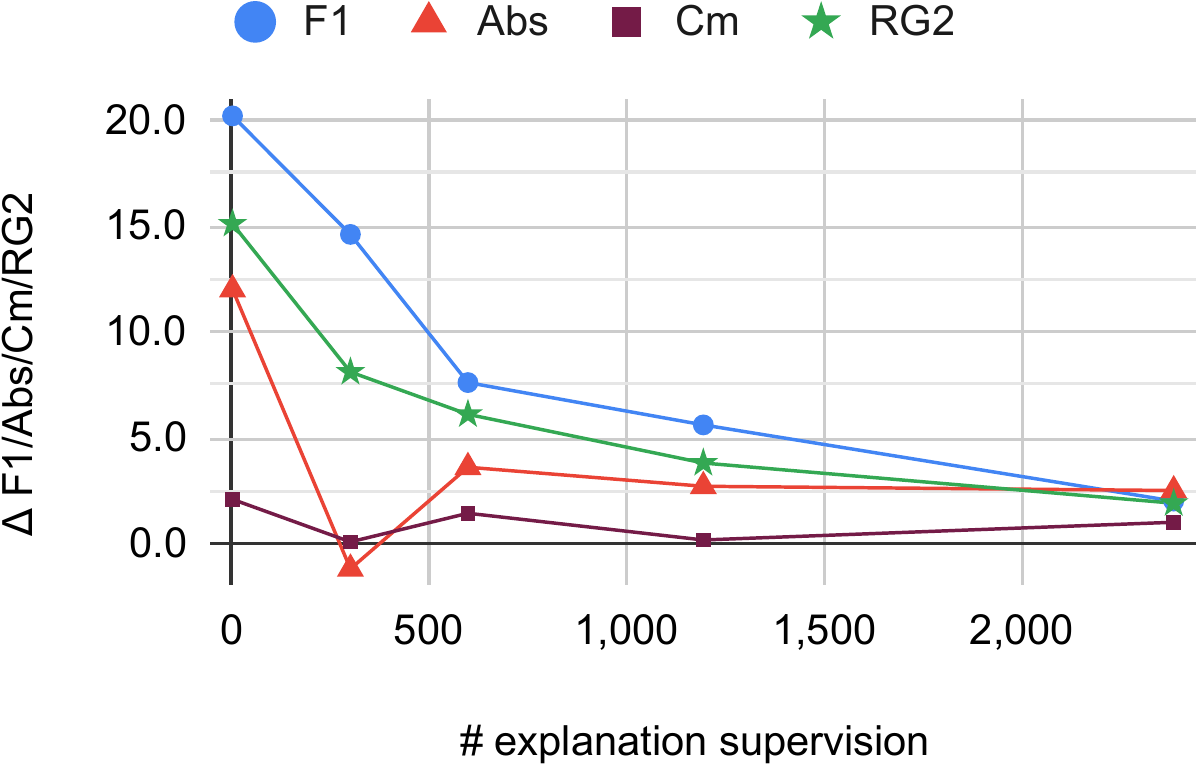}
\caption{
\updated{
Effect of RL.
Y axis indicates the benefit of each evaluation measure from RL (i.e. the difference from \name{}-NoRL to \name{}).
The benefit of RL is more pronounced in low-resource settings.
}
}
\label{fig:size_expl_diff}
\end{figure}

\begin{table}[t]
\centering
\small
\begin{tabular}{llrrrr}\toprule
\textbf{Pretrain?} & \textbf{$L_{\mathrm{ML}}$?} &\textbf{Abs} &\textbf{Cm} &\textbf{RG2$^\dagger$} &\textbf{F1} \\\midrule
SUM &Y &37.5 &11.9 &64.7 &65.3 \\
XG & Y & \textbf{47.0} & \textbf{13.7} & 55.7 & 54.3 \\
SUM,XG &Y &42.6 &12.2 &\textbf{67.4} &\textbf{67.6} \\
\midrule
SUM,XG & &46.3 & 12.7 &52.1 &62.7 \\
\bottomrule
\end{tabular}
\caption{
Ablation of training strategy.
Pretraining on the summarization task plays an important role in generating concise explanations.
Using seq2seq loss $L_{\mathrm{ML}}$ during RL prevents generated explanations from deviating too much from gold explanations.
$\dagger$: evaluated only on 2,541 dev instances annotated with explanations. 
}
\label{tab:ablation_training_strategy}
\end{table}

\begin{table*}[t]
\centering
\small
\begin{tabular}{p{25mm}p{43mm}p{51mm}p{14mm}r}\toprule
\textbf{Insufficiency type} & \textbf{Question} & \textbf{Generated explanation} & \textbf{Gold answer} & \textbf{Freq.} \\
\midrule
No answer span &In which city was this band formed, whose rhythm guitarist featured in "Cupid's Chokehold?" &Cupid's Chokehold is performed by Gym Class Heroes. Fall Out Boy is formed in Wilmette, Illinois. &Chicago &13 \\
Partially missing &Creed features \errorphrase{the boxer} who held what WBC title from 2016 to 2017? &Creed (film) features (\emph{missing: the boxer}) \errorphrase{Tony Bellew}. Tony Bellew held the WBC cruiserweight title from 2016 to 2017. &cruiserweight &8 \\
Bridge fact missing &Where does the descendant of the Red Setter originate? &James Andrew Hanna is known as Red Setter. Scotch Collie originated from the highland regions of Scotland. \emph{Missing: Scotch Collie is the descendant of Red Setter.} &Scotland &3 \\
Fact invented &Which game was released first, Icehouse pieces or Kill Doctor Lucky? &Icehouse pieces was released in \errorphrase{1996} (\emph{correct: 1987}). Kill Doctor Lucky was released in 1996. &Icehouse pieces &1 \\
\midrule
Dataset flaw &Which Walt Disney film was released earlier, The Rescuers or The Muppets? &The Rescuers was released on June 22, 1977. The Muppets was released in 2011. &The Muppets &3 \\
Worker error &Does Lucozade pre-date Hires Root Beer? &Hires Root Beer is introduced in 1876. Lucozade is created in 1927. &no &2 \\
\bottomrule

\end{tabular}
\caption{
Manual analysis of 30 insufficient explanations from \name{}.
}
\label{tab:distr_error}
\end{table*}

\subsection{Analysis}

\subsubsection{Role of explanation supervision}

It is costly to manually annotate QA datasets with abstractive explanations~\cite{inoue-etal-2020-r4c}.
The natural question is then: how much supervision do we need to generate concise explanations?

We pretrain and apply RL, using various sizes of explanation supervision (0, 298, 595, 1190, 2379) and plotted each result in Fig.~\ref{fig:size_expl}.
Due to the cost of human evaluation, we evaluated 100 generated explanations at size 0 and 298 only, and plotted RG2 as a proxy for human-judged sufficiency.

The results indicate that incorporating even 298 explanations has a large impact on both the conciseness of explanations and the QA performance.
Our human-judged sufficiency shows 55.0 for size 0, and 66.0 for size 298.
Even with zero explanation supervision, the explainer still generates concise explanations to some extent.
This indicates that the task of generating abstractive explanations matches with the pretrained summarizer's original task.
Thus, even with such small amounts of data, the \xg{} can learn to produce question-focused summaries that are useful for answering questions.

\updated{
To see the benefit of RL in low-resource settings, we also repeated the same procedure with \name{}-NoRL and plotted how each evaluation measure changes from \name{}-NoRL to \name{} in Fig.~\ref{fig:size_expl_diff}. We observe that the benefit of F1 and RG2 is more pronounced in lower resource settings, which indicates the importance of RL for generating concise explanations.
See \S\ref{sec:lc_norl} in Appendix for the absolute performance of \name{}-NoRL.
}

\subsubsection{Training strategy}
\label{sec:training_str}

\paragraph{Pretraining tasks}
\updated{
We pretrain the \xg{} on the summarization task (SUM) and the explanation generation task (XG) (\S\ref{sec:pretrain}).
To investigate the contribution of each factor, we conduct ablation experiments in Table~\ref{tab:ablation_training_strategy}.
It shows that the summarization task is the most contributing factor: without the pretraining, we obtain more compact explanations, but fatally, they are less similar to the gold explanations and lead to more incorrect answers.
}

\paragraph{Seq2seq loss}
We incur the seq2seq loss ($L_{\mathrm{ML}}$) along with the RL loss (\S\ref{sec:rl}).
To see the effect of this, we conduct ablation experiments in Table~\ref{tab:ablation_training_strategy}.
Without the seq2seq loss, the generated explanations get more compact, but dissimilar to the gold standard explanations.
We speculate that the seq2seq loss is important in keeping the search space of the \xg{} closer to gold explanations.

\subsubsection{Error analysis}

When model's prediction is wrong, we have two possibilities: (A) generated explanations are insufficient, or (B) generated explanations are sufficient, but the QAM fails to find the correct answer.
Table~\ref{tab:conf} indicates that case A is more frequent (69.1\% (38/55)) than case B (30.9\% (17/55)).

We thus randomly sampled and manually analyzed 30 insufficient explanations generated by \name{} in Table~\ref{tab:distr_error}.
First of all, we found that 43.3\% (13/30) of generated explanations have no gold answer spans (`No answer span`).
Among the rest of explanations, the \xg{} successfully mentions important entities, but fails to generate some related information such as entity type (`Partially missing', 26.7\% (8/30)).
We also observed that the \xg{} fails to provide important information bridging two entities such as a family relation (`Bridge fact missing`, 10.0\% (3/30)), and sometimes the \xg{} invents new fact that is not mentioned in the original input paragraph (`Fact invented', 3.3\% (1/30)).

The remaining explanations are wrongly judged as insufficient (16.7\% (5/30)) in 2 cases: (i) crowdworkers' answers were wrongly judged as incorrect due to wrong gold answers (`Dataset flaw`); (ii) the crowdworkers' judgement was wrong, and they are actually sufficient (`Worker error`).

\updated{
The error analysis highlighted that a major source of errors is the explainer failing to include answer spans in generated explanations.
One can possibly enhance our architecture with one more pass: before generating explanations, the QAM predicts candidate answers based on questions and input paragraphs, and feeds them into the explainer.
}

\section{Conclusions}

We have proposed \name{}, an RC system augmented with an abstractive explainer component.
Our experiments have demonstrated that the abstractive explainer can generate more concise explanations than an extractive explainer with limited supervison, while keeping explanations sufficient for QA.

\updated{
One limitation of our work is that the QA module is trained separately from the explainer.
One can jointly optimize the \xg{} and QAM by extending our framework.
Finally, our abstractive explainer explains what facts were used for answering questions, but does not explain the inference process.
It would be an interesting research direction to extend our work by explaining how these facts are combined to arrive at the answer.
}

\section*{Acknowledgements}

\updated{
This work was supported in part by the National Science Foundation under grant No. IIS-1815358 and JST CREST Grant Number JPMJCR20D2, Japan.
We thank the anonymous reviewers for the insightful feedback.
}

\section*{Ethical concerns}

\updated
{
To make fair compensation for Mechanical Turk workers in human evaluation (\S\ref{sec:evalm}), we setup a reward based on a minimum hourly wage in the United States.
Our preliminary experiments show that it takes about one minute to finish one HIT, so 
we rewarded crowdworkers with \$0.15 per HIT.
This amounts to \$9.00 per hour, which is above \$7.25, a minimum wage in the United States.
}

\bibliography{anthology,custom}

\begin{thebibliography}{34}
\expandafter\ifx\csname natexlab\endcsname\relax\def\natexlab#1{#1}\fi

\bibitem[{Baumel et~al.(2016)Baumel, Cohen, and Elhadad}]{baumel2016tcduc}
Tal Baumel, Raphael Cohen, and Michael Elhadad. 2016.
\newblock \href
  {https://www.aaai.org/ocs/index.php/AAAI/AAAI16/paper/download/11939/11989}
  {Topic concentration in query focused summarization datasets}.
\newblock In \emph{AAAI}, pages 2573--2579.

\bibitem[{Bhandari et~al.(2020)Bhandari, Gour, Ashfaq, Liu, and
  Neubig}]{bhandari-etal-2020-evaluating}
Manik Bhandari, Pranav~Narayan Gour, Atabak Ashfaq, Pengfei Liu, and Graham
  Neubig. 2020.
\newblock \href {https://doi.org/10.18653/v1/2020.emnlp-main.751}
  {Re-evaluating evaluation in text summarization}.
\newblock In \emph{Proceedings of the 2020 Conference on Empirical Methods in
  Natural Language Processing (EMNLP)}, pages 9347--9359, Online. Association
  for Computational Linguistics.

\bibitem[{Camburu et~al.(2018)Camburu, Rockt\"{a}schel, Lukasiewicz, and
  Blunsom}]{NEURIPS2018_4c7a167b}
Oana-Maria Camburu, Tim Rockt\"{a}schel, Thomas Lukasiewicz, and Phil Blunsom.
  2018.
\newblock \href
  {https://proceedings.neurips.cc/paper/2018/file/4c7a167bb329bd92580a99ce422d6fa6-Paper.pdf}
  {e-snli: Natural language inference with natural language explanations}.
\newblock In \emph{Advances in Neural Information Processing Systems},
  volume~31. Curran Associates, Inc.

\bibitem[{Dang(2006)}]{dang-2006-duc}
Hoa~Trang Dang. 2006.
\newblock \href {https://www.aclweb.org/anthology/W06-0707} {{DUC} 2005:
  Evaluation of question-focused summarization systems}.
\newblock In \emph{Proceedings of the Workshop on Task-Focused Summarization
  and Question Answering}, pages 48--55, Sydney, Australia. Association for
  Computational Linguistics.

\bibitem[{DeYoung et~al.(2020)DeYoung, Jain, Rajani, Lehman, Xiong, Socher, and
  Wallace}]{deyoung-etal-2020-eraser}
Jay DeYoung, Sarthak Jain, Nazneen~Fatema Rajani, Eric Lehman, Caiming Xiong,
  Richard Socher, and Byron~C. Wallace. 2020.
\newblock \href {https://doi.org/10.18653/v1/2020.acl-main.408} {{ERASER}: {A}
  benchmark to evaluate rationalized {NLP} models}.
\newblock In \emph{Proceedings of the 58th Annual Meeting of the Association
  for Computational Linguistics}, pages 4443--4458, Online. Association for
  Computational Linguistics.

\bibitem[{Glockner et~al.(2020)Glockner, Habernal, and
  Gurevych}]{glockner-etal-2020-think}
Max Glockner, Ivan Habernal, and Iryna Gurevych. 2020.
\newblock \href {https://doi.org/10.18653/v1/2020.findings-emnlp.97} {Why do
  you think that? exploring faithful sentence-level rationales without
  supervision}.
\newblock In \emph{Findings of the Association for Computational Linguistics:
  EMNLP 2020}, pages 1080--1095, Online. Association for Computational
  Linguistics.

\bibitem[{Groeneveld et~al.(2020)Groeneveld, Khot, {Mausam}, and
  Sabharwal}]{groeneveld-etal-2020-simple}
Dirk Groeneveld, Tushar Khot, {Mausam}, and Ashish Sabharwal. 2020.
\newblock \href {https://doi.org/10.18653/v1/2020.emnlp-main.711} {A simple yet
  strong pipeline for {H}otpot{QA}}.
\newblock In \emph{Proceedings of the 2020 Conference on Empirical Methods in
  Natural Language Processing (EMNLP)}, pages 8839--8845, Online. Association
  for Computational Linguistics.

\bibitem[{Guu et~al.(2020)Guu, Lee, Tung, Pasupat, and
  Chang}]{pmlr-v119-guu20a}
Kelvin Guu, Kenton Lee, Zora Tung, Panupong Pasupat, and Mingwei Chang. 2020.
\newblock \href {http://proceedings.mlr.press/v119/guu20a.html} {Retrieval
  augmented language model pre-training}.
\newblock In \emph{Proceedings of the 37th International Conference on Machine
  Learning}, volume 119 of \emph{Proceedings of Machine Learning Research},
  pages 3929--3938. PMLR.

\bibitem[{Hovy et~al.(2013)Hovy, Berg-Kirkpatrick, Vaswani, and
  Hovy}]{hovy-etal-2013-learning}
Dirk Hovy, Taylor Berg-Kirkpatrick, Ashish Vaswani, and Eduard Hovy. 2013.
\newblock \href {https://www.aclweb.org/anthology/N13-1132} {Learning whom to
  trust with {MACE}}.
\newblock In \emph{Proceedings of the 2013 Conference of the North {A}merican
  Chapter of the Association for Computational Linguistics: Human Language
  Technologies}, pages 1120--1130, Atlanta, Georgia. Association for
  Computational Linguistics.

\bibitem[{Inoue et~al.(2020)Inoue, Stenetorp, and Inui}]{inoue-etal-2020-r4c}
Naoya Inoue, Pontus Stenetorp, and Kentaro Inui. 2020.
\newblock \href {https://doi.org/10.18653/v1/2020.acl-main.602} {{R}4{C}: A
  benchmark for evaluating {RC} systems to get the right answer for the right
  reason}.
\newblock In \emph{Proceedings of the 58th Annual Meeting of the Association
  for Computational Linguistics}, pages 6740--6750, Online. Association for
  Computational Linguistics.

\bibitem[{Jacovi and Goldberg(2020)}]{jacovi-goldberg-2020-towards}
Alon Jacovi and Yoav Goldberg. 2020.
\newblock \href {https://doi.org/10.18653/v1/2020.acl-main.386} {Towards
  faithfully interpretable {NLP} systems: How should we define and evaluate
  faithfulness?}
\newblock In \emph{Proceedings of the 58th Annual Meeting of the Association
  for Computational Linguistics}, pages 4198--4205, Online. Association for
  Computational Linguistics.

\bibitem[{Jansen et~al.(2018)Jansen, Wainwright, Marmorstein, and
  Morrison}]{Jansen2018}
Peter~A. Jansen, Elizabeth Wainwright, Steven Marmorstein, and Clayton~T.
  Morrison. 2018.
\newblock \href {https://www.aclweb.org/anthology/L18-1433} {{WorldTree: A
  Corpus of Explanation Graphs for Elementary Science Questions supporting
  Multi-Hop Inference}}.
\newblock In \emph{Proc. of LREC}, pages 2732--2740.

\bibitem[{Khashabi et~al.(2020)Khashabi, Min, Khot, Sabharwal, Tafjord, Clark,
  and Hajishirzi}]{khashabi-etal-2020-unifiedqa}
Daniel Khashabi, Sewon Min, Tushar Khot, Ashish Sabharwal, Oyvind Tafjord,
  Peter Clark, and Hannaneh Hajishirzi. 2020.
\newblock \href {https://doi.org/10.18653/v1/2020.findings-emnlp.171}
  {{UNIFIEDQA}: Crossing format boundaries with a single {QA} system}.
\newblock In \emph{Findings of the Association for Computational Linguistics:
  EMNLP 2020}, pages 1896--1907, Online. Association for Computational
  Linguistics.

\bibitem[{Kim et~al.(2020)Kim, Tang, and Bansal}]{kim-etal-2020-dense}
Hyounghun Kim, Zineng Tang, and Mohit Bansal. 2020.
\newblock \href {https://doi.org/10.18653/v1/2020.acl-main.435} {Dense-caption
  matching and frame-selection gating for temporal localization in
  {V}ideo{QA}}.
\newblock In \emph{Proceedings of the 58th Annual Meeting of the Association
  for Computational Linguistics}, pages 4812--4822, Online. Association for
  Computational Linguistics.

\bibitem[{Kumar and Talukdar(2020)}]{kumar-talukdar-2020-nile}
Sawan Kumar and Partha Talukdar. 2020.
\newblock \href {https://doi.org/10.18653/v1/2020.acl-main.771} {{NILE} :
  Natural language inference with faithful natural language explanations}.
\newblock In \emph{Proceedings of the 58th Annual Meeting of the Association
  for Computational Linguistics}, pages 8730--8742, Online. Association for
  Computational Linguistics.

\bibitem[{Latcinnik and Berant(2020)}]{latcinnik2020explaining}
Veronica Latcinnik and Jonathan Berant. 2020.
\newblock \href {http://arxiv.org/abs/2004.05569} {Explaining question
  answering models through text generation}.
\newblock \emph{arXiv preprint:2004.05569}.

\bibitem[{Lewis et~al.(2020{\natexlab{a}})Lewis, Liu, Goyal, Ghazvininejad,
  Mohamed, Levy, Stoyanov, and Zettlemoyer}]{lewis-etal-2020-bart}
Mike Lewis, Yinhan Liu, Naman Goyal, Marjan Ghazvininejad, Abdelrahman Mohamed,
  Omer Levy, Veselin Stoyanov, and Luke Zettlemoyer. 2020{\natexlab{a}}.
\newblock \href {https://doi.org/10.18653/v1/2020.acl-main.703} {{BART}:
  Denoising sequence-to-sequence pre-training for natural language generation,
  translation, and comprehension}.
\newblock In \emph{Proceedings of the 58th Annual Meeting of the Association
  for Computational Linguistics}, pages 7871--7880, Online. Association for
  Computational Linguistics.

\bibitem[{Lewis et~al.(2020{\natexlab{b}})Lewis, Perez, Piktus, Petroni,
  Karpukhin, Goyal, K\"{u}ttler, Lewis, Yih, Rockt\"{a}schel, Riedel, and
  Kiela}]{NEURIPS2020_6b493230}
Patrick Lewis, Ethan Perez, Aleksandra Piktus, Fabio Petroni, Vladimir
  Karpukhin, Naman Goyal, Heinrich K\"{u}ttler, Mike Lewis, Wen-tau Yih, Tim
  Rockt\"{a}schel, Sebastian Riedel, and Douwe Kiela. 2020{\natexlab{b}}.
\newblock \href
  {https://proceedings.neurips.cc/paper/2020/file/6b493230205f780e1bc26945df7481e5-Paper.pdf}
  {Retrieval-augmented generation for knowledge-intensive nlp tasks}.
\newblock In \emph{Advances in Neural Information Processing Systems},
  volume~33, pages 9459--9474. Curran Associates, Inc.

\bibitem[{Lin(2004)}]{Lin2004}
Chin-Yew Lin. 2004.
\newblock \href {https://www.aclweb.org/anthology/W04-1013} {{ROUGE: A Package
  for Automatic Evaluation of Summaries}}.
\newblock In \emph{Proc. of the Workshop on Text Summarization Branches Out},
  pages 74--81.

\bibitem[{Liu et~al.(2019)Liu, Ott, Goyal, Du, Joshi, Chen, Levy, Lewis,
  Zettlemoyer, and Stoyanov}]{roberta2019}
Yinhan Liu, Myle Ott, Naman Goyal, Jingfei Du, Mandar Joshi, Danqi Chen, Omer
  Levy, Mike Lewis, Luke Zettlemoyer, and Veselin Stoyanov. 2019.
\newblock \href {http://arxiv.org/abs/1907.11692} {Roberta: {A} robustly
  optimized {BERT} pretraining approach}.
\newblock \emph{arXiv preprint:1907.11692}.

\bibitem[{Nema et~al.(2017)Nema, Khapra, Laha, and
  Ravindran}]{nema-etal-2017-diversity}
Preksha Nema, Mitesh~M. Khapra, Anirban Laha, and Balaraman Ravindran. 2017.
\newblock \href {https://doi.org/10.18653/v1/P17-1098} {Diversity driven
  attention model for query-based abstractive summarization}.
\newblock In \emph{Proceedings of the 55th Annual Meeting of the Association
  for Computational Linguistics (Volume 1: Long Papers)}, pages 1063--1072,
  Vancouver, Canada. Association for Computational Linguistics.

\bibitem[{Paranjape et~al.(2020)Paranjape, Joshi, Thickstun, Hajishirzi, and
  Zettlemoyer}]{paranjape-etal-2020-information}
Bhargavi Paranjape, Mandar Joshi, John Thickstun, Hannaneh Hajishirzi, and Luke
  Zettlemoyer. 2020.
\newblock \href {https://doi.org/10.18653/v1/2020.emnlp-main.153} {An
  information bottleneck approach for controlling conciseness in rationale
  extraction}.
\newblock In \emph{Proceedings of the 2020 Conference on Empirical Methods in
  Natural Language Processing (EMNLP)}, pages 1938--1952, Online. Association
  for Computational Linguistics.

\bibitem[{Pasunuru et~al.(2021)Pasunuru, Celikyilmaz, Galley, Xiong, Zhang,
  Bansal, and Gao}]{pasunuru2021data}
Ramakanth Pasunuru, Asli Celikyilmaz, Michel Galley, Chenyan Xiong, Yizhe
  Zhang, Mohit Bansal, and Jianfeng Gao. 2021.
\newblock \href {https://ojs.aaai.org/index.php/AAAI/article/view/17611/17418}
  {Data augmentation for abstractive query-focused multi-document
  summarization}.
\newblock In \emph{AAAI}, pages 13666--13674.

\bibitem[{Raffel et~al.(2020)Raffel, Shazeer, Roberts, Lee, Narang, Matena,
  Zhou, Li, and Liu}]{2020t5}
Colin Raffel, Noam Shazeer, Adam Roberts, Katherine Lee, Sharan Narang, Michael
  Matena, Yanqi Zhou, Wei Li, and Peter~J. Liu. 2020.
\newblock \href {http://jmlr.org/papers/v21/20-074.html} {Exploring the limits
  of transfer learning with a unified text-to-text transformer}.
\newblock \emph{Journal of Machine Learning Research}, 21(140):1--67.

\bibitem[{Rajani et~al.(2019)Rajani, McCann, Xiong, and
  Socher}]{rajani-etal-2019-explain}
Nazneen~Fatema Rajani, Bryan McCann, Caiming Xiong, and Richard Socher. 2019.
\newblock \href {https://doi.org/10.18653/v1/P19-1487} {Explain yourself!
  leveraging language models for commonsense reasoning}.
\newblock In \emph{Proceedings of the 57th Annual Meeting of the Association
  for Computational Linguistics}, pages 4932--4942, Florence, Italy.
  Association for Computational Linguistics.

\bibitem[{Rennie et~al.(2017)Rennie, Marcheret, Mroueh, Ross, and
  Goel}]{Rennie_2017_CVPR}
Steven~J. Rennie, Etienne Marcheret, Youssef Mroueh, Jerret Ross, and Vaibhava
  Goel. 2017.
\newblock \href
  {https://openaccess.thecvf.com/content_cvpr_2017/papers/Rennie_Self-Critical_Sequence_Training_CVPR_2017_paper.pdf}
  {Self-critical sequence training for image captioning}.
\newblock In \emph{Proceedings of the IEEE Conference on Computer Vision and
  Pattern Recognition (CVPR)}.

\bibitem[{Shleifer and Rush(2020)}]{shleifer2020pretrained}
Sam Shleifer and Alexander~M. Rush. 2020.
\newblock \href {http://arxiv.org/abs/2010.13002} {Pre-trained summarization
  distillation}.
\newblock \emph{arXiv preprint:2010.13002}.

\bibitem[{Thayaparan et~al.(2020)Thayaparan, Valentino, and
  Freitas}]{thayaparan2020survey}
Mokanarangan Thayaparan, Marco Valentino, and André Freitas. 2020.
\newblock \href {http://arxiv.org/abs/2010.00389} {A survey on explainability
  in machine reading comprehension}.
\newblock \emph{arXiv preprint:2010.00389}.

\bibitem[{Trivedi et~al.(2020)Trivedi, Balasubramanian, Khot, and
  Sabharwal}]{trivedi-etal-2020-multihop}
Harsh Trivedi, Niranjan Balasubramanian, Tushar Khot, and Ashish Sabharwal.
  2020.
\newblock \href {https://doi.org/10.18653/v1/2020.emnlp-main.712} {Is multihop
  {QA} in {DiRe} condition? measuring and reducing disconnected reasoning}.
\newblock In \emph{Proceedings of the 2020 Conference on Empirical Methods in
  Natural Language Processing (EMNLP)}, pages 8846--8863, Online. Association
  for Computational Linguistics.

\bibitem[{Tu et~al.(2020)Tu, Huang, Wang, Huang, He, and Zhou}]{tu2020sae}
Ming Tu, Kevin Huang, Guangtao Wang, Jing Huang, Xiaodong He, and Bowen Zhou.
  2020.
\newblock \href {https://aaai.org/ojs/index.php/AAAI/article/view/6441}
  {Select, answer and explain: Interpretable multi-hop reading comprehension
  over multiple documents}.
\newblock In \emph{{AAAI}}, pages 9073--9080.

\bibitem[{Warstadt et~al.(2018)Warstadt, Singh, and
  Bowman}]{warstadt2018neural}
Alex Warstadt, Amanpreet Singh, and Samuel~R Bowman. 2018.
\newblock \href {http://arxiv.org/abs/1805.12471} {Neural network acceptability
  judgments}.
\newblock \emph{arXiv preprint:1805.12471}.

\bibitem[{Wiegreffe and Marasović(2021)}]{wiegreffe2021teach}
Sarah Wiegreffe and Ana Marasović. 2021.
\newblock \href {http://arxiv.org/abs/2102.12060} {Teach me to explain: A
  review of datasets for explainable nlp}.
\newblock \emph{arXiv preprint:2102.12060}.

\bibitem[{Yadav et~al.(2020)Yadav, Bethard, and
  Surdeanu}]{yadav-etal-2020-unsupervised}
Vikas Yadav, Steven Bethard, and Mihai Surdeanu. 2020.
\newblock \href {https://doi.org/10.18653/v1/2020.acl-main.414} {Unsupervised
  alignment-based iterative evidence retrieval for multi-hop question
  answering}.
\newblock In \emph{Proceedings of the 58th Annual Meeting of the Association
  for Computational Linguistics}, pages 4514--4525, Online. Association for
  Computational Linguistics.

\bibitem[{Yang et~al.(2018)Yang, Qi, Zhang, Bengio, Cohen, Salakhutdinov, and
  Manning}]{yang-etal-2018-hotpotqa}
Zhilin Yang, Peng Qi, Saizheng Zhang, Yoshua Bengio, William Cohen, Ruslan
  Salakhutdinov, and Christopher~D. Manning. 2018.
\newblock \href {https://doi.org/10.18653/v1/D18-1259} {{H}otpot{QA}: A dataset
  for diverse, explainable multi-hop question answering}.
\newblock In \emph{Proceedings of the 2018 Conference on Empirical Methods in
  Natural Language Processing}, pages 2369--2380, Brussels, Belgium.
  Association for Computational Linguistics.

\end{thebibliography}
\bibliographystyle{acl_natbib}

\newpage
\appendix

\section{Training detail}
\label{sec:training detail}

For all experiments, we used public implementations from huggingface's transformers library available at \url{https://huggingface.co/}.
We used \texttt{roberta-large} for the paragraph ranker, \texttt{distilbart-cnn-12-6} for \xg{}, and \texttt{unifiedqa-t5-base} for UnifiedQA-base.

For Reinforcement Learning, we used AdamW with the learning rate of 2e-6 and the batch size of 8.
We clipped the minimum reward to -0.001.
For sampling, we used a temperature of 0.4.
To prevent overfitting, we used early stopping with a patience of 5.
Specifically, we monitor the Answer F1 on the validation set every 4096 training steps and stopped training if the best F1 is not updated for five times.
The RL training took 10h31m on a single GPU (DGXA-100).

For pretraining the \xg{}, we used AdamW with the learning rate of 8e-6 and the batch size of 16.
In all experiments, we used a linear learning rate scheduler with 10\% warm up and trained the models with 5 epochs.
For the learning curve, we monitored the Answer F1 every 128 steps for size 298, 256 steps for size 595, 512 steps for size 1,190 \& 2,379 and used early stopping with a patience of 5.
We used 512 as a maximum length of input subwords for both the \xg{} and QAM.
We used 256 as a maximum length of generation outputs for the \xg{}.
We used greedy decoding for both the \xg{} and QAM.

\section{Experiments with stronger QA model}
\label{sec:largeqa}

\updated{
We conducted additional analysis with SAE-large~\cite{tu2020sae}, one of the large QA models top-ranked at the leaderboard.\footnote{\url{https://hotpotqa.github.io/}}
We downloaded a publicly available pretrained model\footnote{\url{https://github.com/JD-AI-Research-Silicon-Valley/SAE}} and ran the exactly same experiments in Table~\ref{tab:upper_bound}, \ref{tab:main}, and \ref{tab:conf}, where we used SAE-large as the QAM \emph{at test time only}.
Note that during training, we used UnifiedQA-base \emph{not} finetuned on HotpotQA (see \S\ref{sec:setup} for further details).

The results are shown in Table~\ref{tab:largeqa_main} and Table~\ref{tab:largeqa_conf}.
Overall, they show the same trend as Table~\ref{tab:upper_bound}, \ref{tab:main}, and \ref{tab:conf}: (i) gold abstractive explanations yields higher F1; (ii) \name{} achieved better XF1 than the extractive baseline; and (iii) there are more correct answers led by insufficient explanations in the extractive baseline.
}

\begin{table}[h]
\centering
\small
\begin{tabular}{lrrrrrr}\toprule
\textbf{Model} & \textbf{F1} & \textbf{XF1}$^\ddagger$ \\
\midrule
Gold SF$^\dagger$ & 80.1 & - \\
Gold SF & 77.7 & - \\
Gold XP$^\dagger$ & \textbf{84.4} & - \\
\midrule
QAM w/o AX & 70.7 & - \\
\baselinename{} & \textbf{71.5} & 59.4 \\
\name{}-NoRL & 64.9 & 58.5 \\
\name{} & 66.8 & \textbf{60.4} \\
\bottomrule
\end{tabular}
\caption{Larger QA models on HotpotQA (HQ) dev set. $\dagger$: evaluated only on 2,541 dev instances annotated with explanations. $\ddagger$: evaluated on 200 instances with human-judged sufficiency.}
\label{tab:largeqa_main}
\end{table}

\begin{table}[h]
\centering
\small

\begin{tabular}{c}
\begin{minipage}{0.5\linewidth}
\begin{tabular}{lrrr}\toprule
&Correct &Wrong \\\midrule
Suf. & \goodhi{122} & 14 \\
Insuf. & \badhi{28} & 30 \\
\midrule
Total & 150 & 44 \\
\bottomrule
\end{tabular}
\subcaption{\baselinename{} (baseline)}
\end{minipage}

\begin{minipage}{0.5\linewidth}
\begin{tabular}{lrrr}\toprule
&Correct &Wrong \\\midrule
Suf. & \goodhi{120} & 18 \\
Insuf. & \badhi{14} & 41 \\
\midrule
Total & 134 & 59 \\
\bottomrule
\end{tabular}
\subcaption{\name{}}
\end{minipage}
\end{tabular}

\caption{
Sufficiency-Answer correctness matrix.
\name{} gets more correct answers with sufficient explanations (120/134=90\%) than \baselinename{} (122/150=81\%).
}
\label{tab:largeqa_conf}
\end{table}

\section{Learning curve of \name{}-NoRL}
\label{sec:lc_norl}

\begin{figure}[h]
\includegraphics[width=\linewidth]{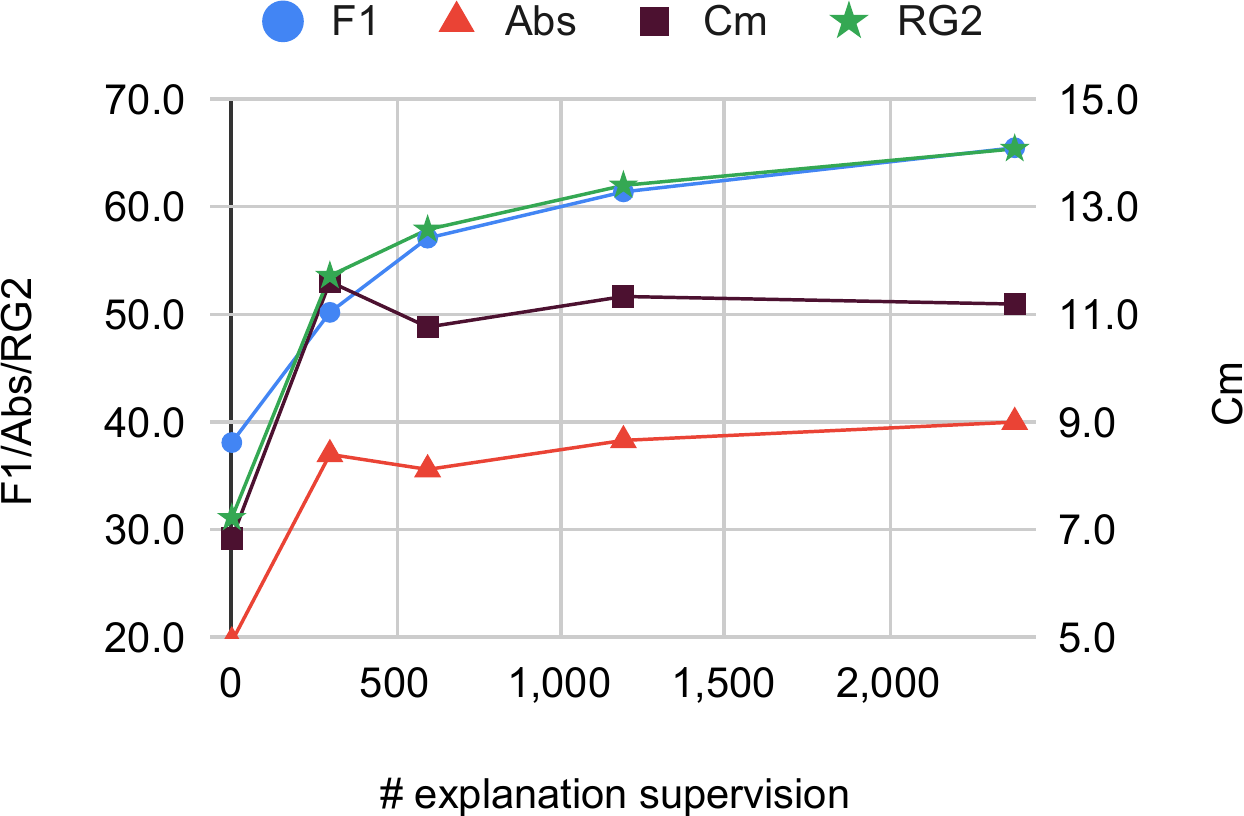}
\caption{
Size of explanation supervision v.s. QA performance and conciseness for \name{}-NoRL.
}
\label{fig:size_expl_norl}
\end{figure}

\updated{
To see the effectiveness of RL in low-resource settings, we investigated the performance change from \name{}-NoRL to \name{} in Fig.~\ref{fig:size_expl_diff}.
Here we plot the absolute performance of \name{}-NoRL in Fig.~\ref{fig:size_expl_norl}.
}

\section{Human evaluation}
\label{sec:humaneval}

We use Mechanical Turk as a crowdsourcing platform for human evaluation.
We hired five annotators per Human Intelligence Task (HIT) and rewarded them with \$0.15.
Our preliminary experiments show that it takes about one minute to finish one HIT, so it is \$9.00 per hour, which is above \$7.25, a minimum wage in the United States.
To ensure the quality of annotations, we used crowdworkers with $\geq 5,000$ HITs experiences and $\geq 99\%$ approval rates.
Among them, we manually find the pool of high-quality workers and used the same pool throughout the experiments.

\begin{figure*}[t!]
\includegraphics[width=\linewidth]{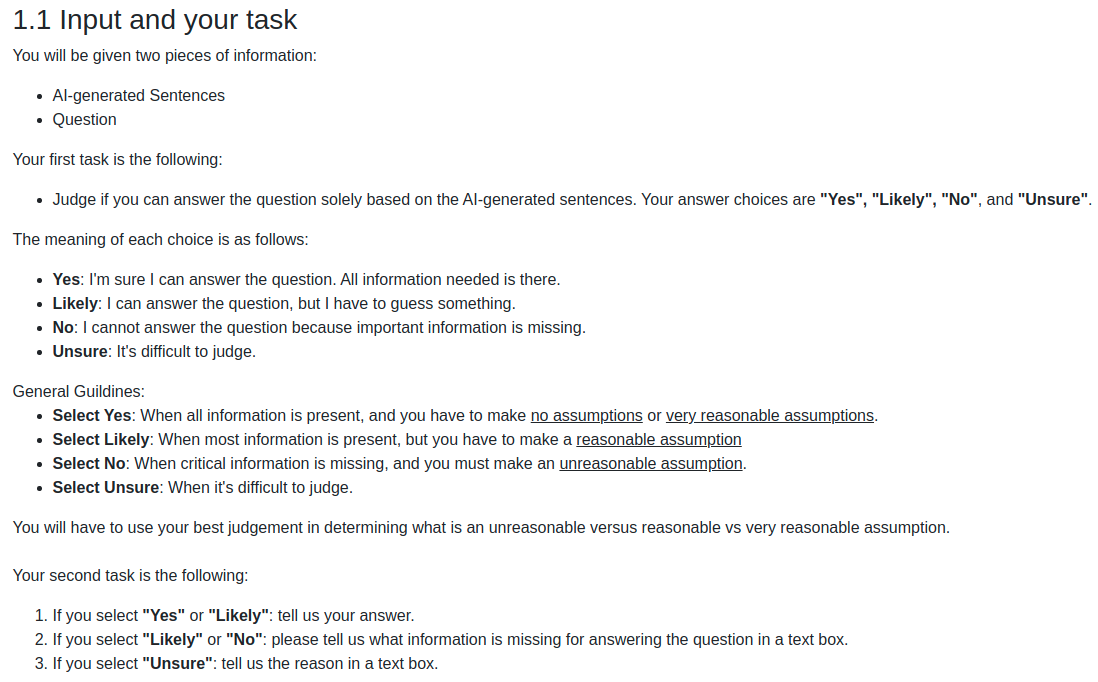}
\caption{
Instruction for crowdworkers (general guidelines).
}
\label{fig:eval_inst_1}
\end{figure*}

\begin{figure*}[t!]
\includegraphics[width=\linewidth]{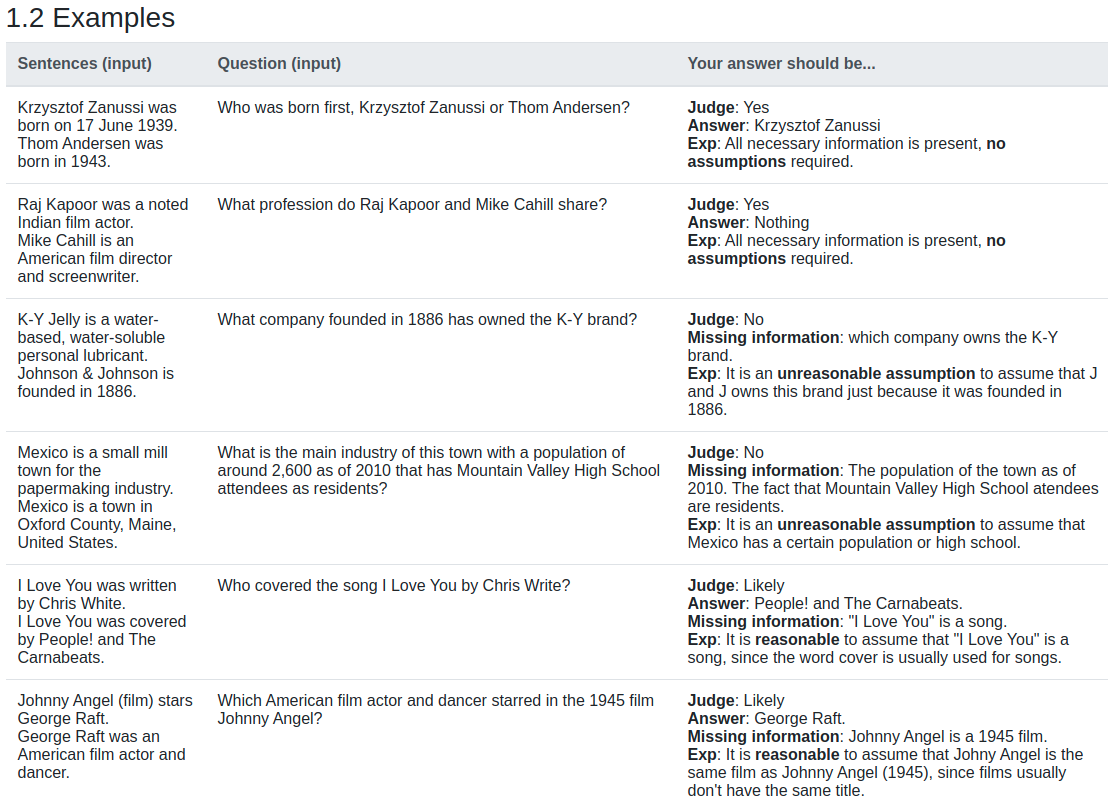}
\caption{
Instruction for crowdworkers (examples).
}
\label{fig:eval_inst_2}
\end{figure*}

\begin{figure*}[t!]
\includegraphics[width=\linewidth]{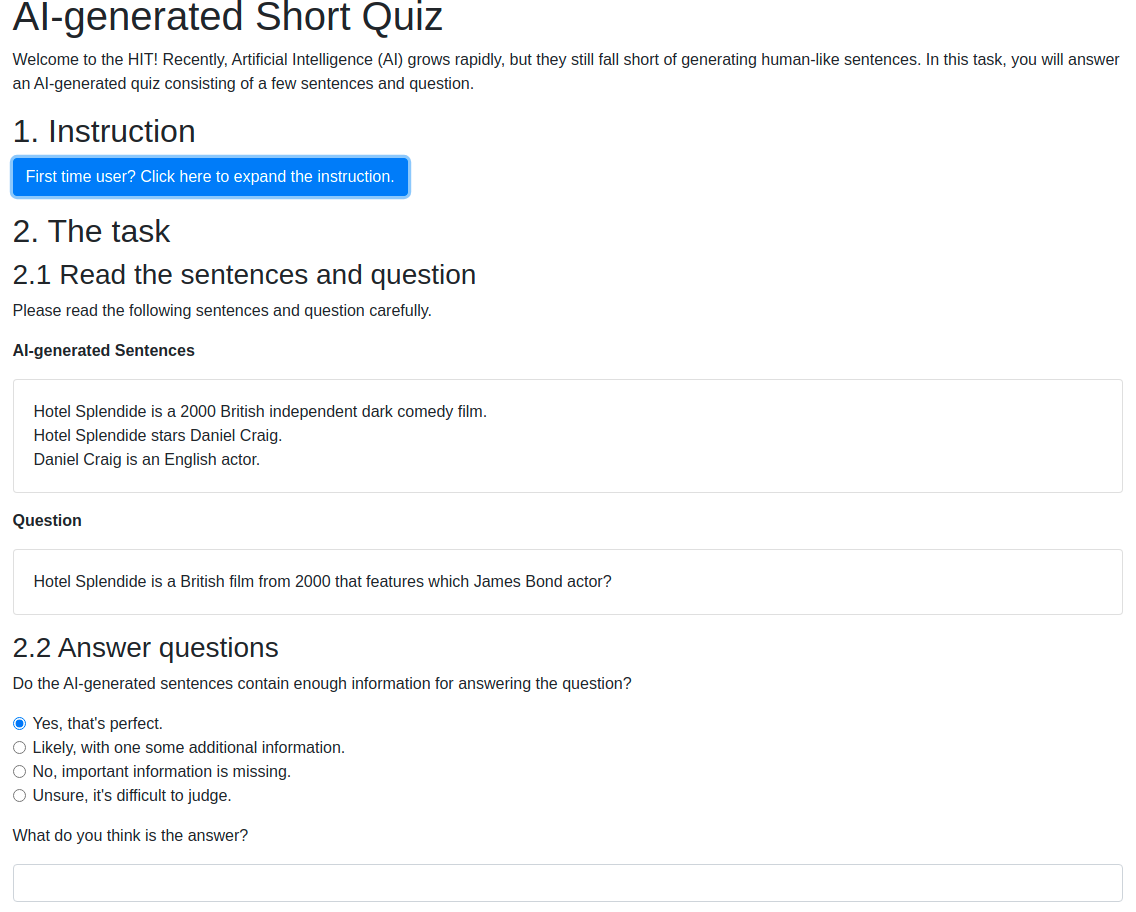}
\caption{
Crowdsourcing interface for sufficiency evaluation.
}
\label{fig:eval_interface}
\end{figure*}

The instruction to crowdworkers is shown in Fig.~\ref{fig:eval_inst_1} and  Fig.~\ref{fig:eval_inst_2}, and the task interface is shown in Fig~\ref{fig:eval_interface}.

\section{Example of generated explanations with full inputs}
\label{sec:fulloutputs}

Examples of generated explanations and predicted answers along with their full input paragraphs retrieved by the paragraph ranker are shown in Table~\ref{tab:full_example_1}, Table~\ref{tab:full_example_2} and Table~\ref{tab:full_example_3}.

\begin{table*}[t]
\centering
\small
\begin{tabular}{p{20mm}p{70mm}p{35mm}p{10mm}}\toprule
\textbf{Question} & \textbf{Input paragraphs from the ranker} & \textbf{Generated explanation} & \textbf{Predicted answer} \\
\midrule
Charlie Rowe plays Billy Costa in a film based on what novel?
&
\importantent{[P1]} Charles John Rowe is an English actor.
His film roles include Young Tommy in "Never Let Me Go", James in "The Boat That Rocked", Billy Costa in "The Golden Compass" , Peter in the SyFy/Sky Movies Peter Pan prequel "Neverland", and recently played Leo Roth on the Fox medical comedy-drama series "Red Band Society" alongside Octavia Spencer and Dave Annable.
The Golden Compass is a 2007 British-American fantasy adventure film based on "Northern Lights", the first novel in Philip Pullman's trilogy "His Dark Materials".
Written and directed by Chris Weitz, it stars Nicole Kidman, Dakota Blue Richards, Daniel Craig, Sam Elliott, Eva Green, and Ian McKellen.
The project was announced in February 2002, but difficulties over the script and the selection of a director caused significant delays.
At US\$180 million, it was one of New Line Cinema's most expensive projects ever, and its disappointing results in the USA contributed to New Line's February 2008 restructuring.
Charlie St.
\importantent{[P2]} Cloud is a 2010 American drama film based on Ben Sherwood's best-selling novel, "The Death and Life of Charlie St.
Cloud" published in 2004 by Bantam Books.
The film is directed by Burr Steers and stars Zac Efron and Amanda Crew.
The story is of Charlie St.
Cloud's choice between keeping a promise he made to his brother, who died in a car accident, or going after the girl he loves.
In some markets the film used the complete title of the book.
\importantent{[P3]} The Golden Compass is a 2007 British-American fantasy adventure film based on "Northern Lights", the first novel in Philip Pullman's trilogy "His Dark Materials".
Written and directed by Chris Weitz, it stars Nicole Kidman, Dakota Blue Richards, Daniel Craig, Sam Elliott, Eva Green, and Ian McKellen.
The project was announced in February 2002, but difficulties over the script and the selection of a director caused significant delays.
At US\$180 million, it was one of New Line Cinema's most expensive projects ever, and its disappointing results in the USA contributed to New Line's February 2008 restructuring.
Charles John Rowe is an English actor.
His film roles include Young Tommy in "Never Let Me Go", James in "The Boat That Rocked", Billy Costa in "The Golden Compass" , Peter in the SyFy/Sky Movies Peter Pan prequel "Neverland", and recently played Leo Roth on the Fox medical comedy-drama series "Red Band Society" alongside Octavia Spencer and Dave Annable.
400 Boys is a Canadian/Chinese action-fantasy thriller film directed, written and produced by Alastair Paton and starring Jodelle Ferland, Li Bingbing, Maria Valverde and Charlie Rowe.	Northern Lights	"Northern Lights"	Charlie Rowe plays Billy Costa in The Golden Compass.
The Golden Compass is based on Northern Lights.
&
Charlie Rowe plays Billy Costa in The Golden Compass.
The Golden Compass is based on Northern Lights.
&
Northern Lights
\\
\bottomrule
\end{tabular}
\caption{
Full input and output example (1).
}
\label{tab:full_example_1}
\end{table*}

\begin{table*}[t]
\centering
\small
\begin{tabular}{p{20mm}p{70mm}p{35mm}p{10mm}}\toprule
\textbf{Question} & \textbf{Input paragraphs from the ranker} & \textbf{Generated explanation} & \textbf{Predicted answer} \\
\midrule
Who was born first, Krzysztof Zanussi or Thom Andersen?
&
\importantent{[P1]} Krzysztof Zanussi, (born 17 June 1939) is a Polish film and theatre director, producer and screenwriter.
He is a professor of European film at the European Graduate School in Saas-Fee, Switzerland where he conducts a summer workshop.
He is also a professor at the Silesian University in Katowice.
\importantent{[P2]} Thom Andersen (born 1943, Chicago) is a filmmaker, film critic and teacher.
\importantent{[P3]} Weronika Anna Rosati (] ; born 9 January 1984) is a Polish actress and a member of European Film Academy.
She began her acting career in Polish soaps.
In 2005, she starred as Dżemma in her first theatrical feature film "Pitbull".
A year later, she launched her international career with a small uncredited role in "Inland Empire" directed by David Lynch.
Since then she has appeared in many critically acclaimed Polish and international productions.
In 2013, she has received her first Polish Academy Award nomination for Best Actress for her role in "Obława" (2012).
A year later, she starred alongside Agnieszka Grochowska in "Obce ciało" directed by Krzysztof Zanussi.
She also had a recurring role in the HBO TV series "Luck" (2012).
&
Krzysztof Zanussi is born on 17 June 1939.
Thom Andersen is born on 1943.
&
Krzysztof Zanussi
\\
\bottomrule
\end{tabular}
\caption{
Full input and output example (2).
}
\label{tab:full_example_2}
\end{table*}

\begin{table*}[t]
\centering
\small
\begin{tabular}{p{20mm}p{70mm}p{35mm}p{10mm}}\toprule
\textbf{Question} & \textbf{Input paragraphs from the ranker} & \textbf{Generated explanation} & \textbf{Predicted answer} \\
\midrule
Kalle Mäkinen is nicknamed after a Spanish football that captains what two teams?
&
\importantent{[P1]} Kalle Mäkinen (born 1 February 1989) is a Finnish footballer who last played for the Finnish Veikkausliiga club Maskun Palloseura.
He was nicknamed "Sergio" because his style of play was similar to that of Spanish fullback Sergio Ramos.
\importantent{[P2]} Sergio Ramos García (] ; born 30 March 1986) is a Spanish professional footballer who plays for and captains both Real Madrid and the Spain national team.
Primarily a central defender, he can also play as a right back.
\importantent{[P3]} The Primera División, commonly known as La Liga and as La Liga Santander for sponsorship reasons with Santander, is the top professional association football division of the Spanish football league system.
Administrated by the Liga de Fútbol Profesional (LFP), La Liga is contested by 20 teams, with the three lowest-placed teams relegated to the Segunda División and replaced by the top two teams in that division plus the winner of a play-off.
&
Kalle Mäkinen is nicknamed Sergio Ramos.
Sergio Ramos captains both Real Madrid and the Spain national team.
&
Real Madrid and the Spain national team
\\
\bottomrule
\end{tabular}
\caption{
Full input and output example (3).
}
\label{tab:full_example_3}
\end{table*}

\end{document}